\documentclass{article}

\usepackage{amsmath}
\usepackage{amssymb}
\usepackage{graphicx}
\usepackage{hyperref}
\usepackage{url}
\usepackage{tikz}
\usetikzlibrary{shadings}
\usepackage{iclr2024/iclr2024_conference}

\usepackage{amsmath,amsfonts,bm}

\def\eqref#1{equation~\ref{#1}}

\def\1{\bm{1}}

\DeclareMathAlphabet{\mathsfit}{\encodingdefault}{\sfdefault}{m}{sl}
\SetMathAlphabet{\mathsfit}{bold}{\encodingdefault}{\sfdefault}{bx}{n}

\title{Precise Bayesian Neural Networks}

\author{Carlos Stein Brito \\
NightCity Labs, Lisbon, Portugal. \\
\texttt{carlos.stein@nightcitylabs.ai}}

\iclrfinalcopy

\begin{document}

\maketitle

\begin{abstract}
Despite its long history, Bayesian neural networks (BNNs) and variational training remain underused in practice: standard Gaussian posteriors misalign with network geometry, KL terms can be brittle in high dimensions, and implementations often add complexity without reliably improving uncertainty. We revisit the problem through the lens of normalization. Because normalization layers neutralize the influence of weight magnitude, we model uncertainty \emph{only in weight directions} using a von Mises-Fisher posterior on the unit sphere. High-dimensional geometry then yields a single, interpretable scalar per layer--the effective post-normalization noise $\sigma_{\mathrm{eff}}$--that (i) corresponds to simple additive Gaussian noise in the forward pass and (ii) admits a compact, dimension-aware KL in closed form. We derive accurate, closed-form approximations linking concentration $\kappa$ to activation variance and to $\sigma_{\mathrm{eff}}$ across regimes, producing a lightweight, implementation-ready variational unit that fits modern normalized architectures and improves calibration without sacrificing accuracy. This dimension awareness is critical for stable optimization in high dimensions. In short, by aligning the variational posterior with the network's intrinsic geometry, BNNs can be simultaneously principled, practical, and precise.
\end{abstract}

\section{Introduction}
\label{sec:intro}
BNNs promise calibrated predictions but are rarely used in deployed systems. Standard VI for BNNs 
\citep{Bishop2006,Graves2011practicalvi,Jordan1999,Kingma2014autoencoding} commonly adopts factorized Gaussian posteriors \citep{Kingma2015variational,Molchanov2017variational} or MC Dropout \citep{Gal2016dropout}. In practice, these choices often misalign with modern network geometry, yield brittle KL terms in high dimensions, and add engineering complexity without reliably improving calibration \citep{Guo2017calibration,Mukhoti2020calibrating,Izmailov2021bayesian}.

In architectures with pervasive normalization, a layer’s function depends mainly on the \emph{direction} of its weights. Batch Normalization (BN) makes pre-activation scale largely irrelevant \citep{Ioffe2015batchnorm}; Weight Normalization separates direction and scale \citep{Salimans2016weightnorm}; and scale-symmetry analyses formalize the role of directions in ReLU networks \citep{Meng2019gsgd,Neyshabur2015normbased}. Modeling a full Euclidean posterior over both radius and direction is therefore unnecessary and can be fragile in high dimension. Furthermore, dimension-agnostic regularizers can lead to poorly scaled KL terms in wide layers, hampering stability and calibration.

We instead place the approximate posterior on the unit sphere: directions follow a von Mises-Fisher (vMF) distribution with a uniform spherical prior \citep{MardiaJupp2000,Banerjee2005clustering,davidson2018hyperspherical}. Prior work considered radial-directional decompositions or spherical constraints \citep{oh2020radial,farquhar2020radial,ghoshal2021hyperspherical}; here we provide a formulation matched to normalized networks that yields a \emph{single scalar per layer} controlling both injected noise and the KL term.

After normalization, additive Gaussian noise with standard deviation $\sigma_{\mathrm{eff}}$ matches the effect (in expectation) of sampling a weight direction from vMF with concentration $\kappa$. High-dimensional geometry gives compact relations between $\kappa$, pre-activation variance, and $\sigma_{\mathrm{eff}}$, and leads to a closed-form, dimension-aware KL expressed directly in terms of $\sigma_{\mathrm{eff}}$ and input dimension $D$ (matching vMF KL asymptotics) \citep{Sra2016}. The KL scales with the layer's input dimensionality $D$.

\textbf{Contributions.} (i) A geometry-aligned directional posterior for weights (vMF with uniform prior) tailored to normalized networks. (ii) A compact activation-variance approximation $\sigma_u^2(\kappa) \approx D/(\kappa+D)$. (iii) An effective post-normalization noise parameter $\sigma_{\mathrm{eff}} = \sigma_u/A_D(\kappa)$ with a unified, closed-form KL $\mathrm{KL}_{\mathrm{approx}}(\sigma_{\mathrm{eff}},D)$. (iv) A practical layer design using one scalar per normalized layer that integrates into standard training and improves calibration on CIFAR-10 with VGG16+BN.

The next sections present the geometry, the variance and KL approximations, and experiments.

\begin{figure}[t]
\centering
\begin{minipage}[t]{0.49\textwidth}
  \centering
  \includegraphics[width=\linewidth]{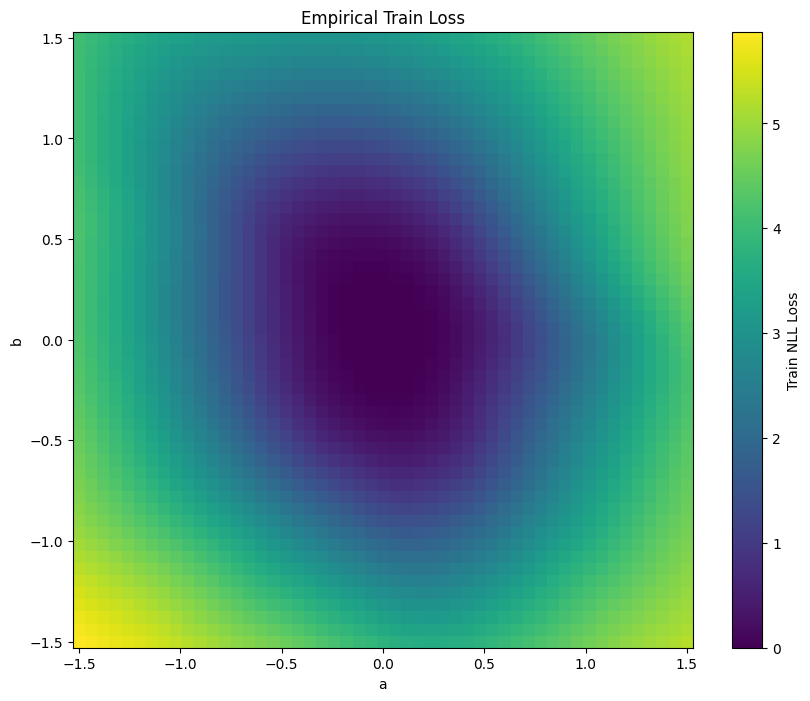}
  \vspace{0.2em}
  \small Empirical 2D train-loss around a trained layer.
\end{minipage}
\hfill
\begin{minipage}[t]{0.49\textwidth}
  \centering
  \includegraphics[width=\linewidth]{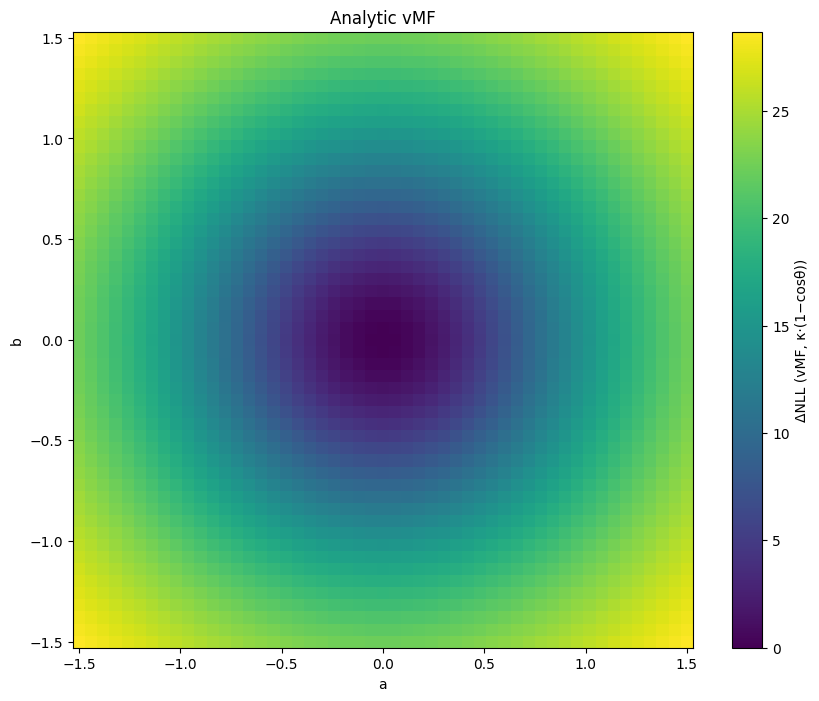}
  \vspace{0.2em}
  \small Analytic vMF directional model (same coordinates).
\end{minipage}
\caption{Motivational comparison. The empirical landscape is bowl-shaped with near-radial level sets in the 2D tangent plane, which is naturally captured by a directional (vMF) model rather than Euclidean distance.}
\label{fig:intro_motivational}
\end{figure}

\section{Related Work}
\label{sec:related}
\textbf{BNNs and variational inference.} Bayesian neural networks aim for calibrated predictive distributions by maintaining posterior uncertainty over parameters \citep{Bishop2006,Graves2011practicalvi}. Variational inference \citep{Jordan1999,Kingma2014autoencoding} with factorized Gaussian posteriors is widely used \citep{Kingma2015variational,Molchanov2017variational}, and MC Dropout offers a heuristic alternative \citep{Gal2016dropout}. Empirically, strong calibration remains challenging \citep{Guo2017calibration,Izmailov2021bayesian}.

\textbf{Geometry and directional statistics.} Several works exploit spherical structure: radial-directional decompositions in weight space \citep{oh2020radial,farquhar2020radial}, hyperspherical constraints for weights \citep{ghoshal2021hyperspherical}, and vMF-based variational models in latent spaces \citep{davidson2018hyperspherical}. Directional statistics on $\mathbb S^{D-1}$ provide the natural family for unit-norm variables \citep{MardiaJupp2000,Banerjee2005clustering}. Our approach differs by placing the \emph{weight posterior} on the sphere and deriving a closed-form, dimension-aware KL in the optimization variable $\sigma_{\mathrm{eff}}$ (matching vMF KL asymptotics \citep{Sra2016}).

\textbf{Normalization and scale invariance.} Batch Normalization \citep{Ioffe2015batchnorm} and Weight Normalization \citep{Salimans2016weightnorm} separate direction from magnitude, and scale-symmetry analyses highlight direction as the informative degree of freedom in ReLU networks \citep{Meng2019gsgd,Neyshabur2015normbased}. We leverage this to model \emph{directional} uncertainty only, yielding a single scalar per layer that controls post-normalization noise and a principled KL.

\textbf{Positioning.} Relative to Gaussian mean-field VI and variational dropout, our method replaces radial weight uncertainty with a sphere-aligned posterior and uses an analytic, dimension-aware KL. Compared to spherical constraints and prior vMF uses, we provide a direct link from concentration to effective post-normalization noise and a closed-form training objective.

\textbf{Dimension-aware scaling.} Our emphasis on dimension-dependent regularization resonates with insights from scaling and initialization theory (e.g., $\mu$-parametrization \citep{Yang2022tensorprogramsV}), which similarly tie stable behavior and effective capacity to layer dimensionality.

\section{Geometric Preliminaries and Normalization Invariance}
\label{sec:geometry}
Let $w\in\mathbb{R}^D$ be a weight vector, $x\in\mathbb{R}^D$ an input, and $u=w^\top x$ the pre-activation. Write the unit direction $\tilde w:=w/\|w\|$ and the unit sphere $\mathbb S^{D-1}$. We focus on architectures with Batch Normalization (BN) or equivalent normalization between linear/convolutional layers and nonlinearities.

\subsection{Normalization removes scale}
Normalization layers without affine parameters (e.g., Batch Normalization) normalize $u$ by subtracting its batch mean and dividing by its standard deviation. For any $\alpha>0$, $\mu_{\alpha u}=\alpha\mu_u$ and $\sigma_{\alpha u}=\alpha\sigma_u$, hence
\begin{equation}
\mathrm{BN}(\alpha u)=\frac{\alpha u-\alpha\mu_u}{\alpha\sigma_u}=\mathrm{BN}(u)\quad\Rightarrow\quad \mathrm{BN}(w^\top x)=\mathrm{BN}(\tilde w^\top x).
\end{equation}
Thus, downstream computation depends primarily on weight \emph{direction} rather than magnitude \citep{Ioffe2015batchnorm,Salimans2016weightnorm,Meng2019gsgd,Neyshabur2015normbased}.

\subsection{Directional posteriors on the sphere}
We model uncertainty on $\mathbb S^{D-1}$ with a von Mises-Fisher (vMF) distribution with mean direction $\mu\in\mathbb S^{D-1}$ and concentration $\kappa\ge0$:
\begin{equation}
q(w\mid\mu,\kappa)=C_D(\kappa)\,\exp\bigl(\kappa\,\mu^\top w\bigr),\qquad w\in\mathbb S^{D-1},
\end{equation}
where $C_D(\kappa)$ is the normalizer and the mean resultant length is $A_D(\kappa):=I_{D/2}(\kappa)/I_{D/2-1}(\kappa)$ \citep{MardiaJupp2000,Banerjee2005clustering}. We use a uniform spherical prior.

\subsection{From directional uncertainty to activation variance}
Let $\mathrm{Cov}_q(w)$ be the covariance of $w\sim\mathrm{vMF}(\mu,\kappa)$. By symmetry,
\begin{equation}
\mathrm{Cov}_q(w)=\sigma_{\parallel}^2\,\mu\mu^\top+\sigma_{\perp}^2\,(I-\mu\mu^\top),\quad \Rightarrow\quad \sigma_u^2:=\mathrm{Var}_q(w^\top x)=\sigma_{\parallel}^2(\mu^\top x)^2+\sigma_{\perp}^2\bigl(\|x\|^2-(\mu^\top x)^2\bigr).
\end{equation}
For inputs that are approximately isotropic after normalization with typical norm $\|x\|^2\approx D$, we use the compact, dimension-aware approximation
\begin{equation}
\sigma_u^2(\kappa)\;\approx\;\frac{D}{\kappa+D},
\end{equation}
which recovers saturation as $\kappa\to0$ and the $\propto1/\kappa$ decay as $\kappa\to\infty$.

\subsection{Effective post-normalization noise}
The mean direction shrinks by $A_D(\kappa)$; measuring noise relative to this mean leads to
\begin{equation}
\sigma_{\mathrm{eff}}(\kappa):=\frac{\sigma_u(\kappa)}{A_D(\kappa)}.
\end{equation}
In practice, injecting additive Gaussian noise of standard deviation $\sigma_{\mathrm{eff}}$ after normalization matches the effect (in expectation) of sampling directions from vMF. Subsequent sections use $\sigma_{\mathrm{eff}}$ as the optimized variable and employ a closed-form, dimension-aware KL written directly in terms of $(\sigma_{\mathrm{eff}},D)$.

\begin{figure}[t]
\centering
\begin{tikzpicture}[scale=1]
  \shade[ball color=blue!10,opacity=0.5] (0,0) circle (2);

  \draw[->,thick,blue!60] (0,0) -- (0.95,0.95) node[anchor=west]{$\mu$};

  \draw[->,gray] (0,0) -- (0.95,1.17);
  \draw[->,gray] (0,0) -- (0.8,1.3);
  \draw[->,gray] (0,0) -- (1.05,0.85);

  \draw[blue!60,thick,fill=blue!60,fill opacity=0.08] (0.95,0.95) ellipse [x radius=0.35, y radius=0.5, rotate=45];
  \node[blue!60,anchor=west] at (0.95,0.3) {$\kappa$};

  \draw (5,0) circle (2);

  \draw[->,thick,blue!60] (5,0) -- (5,2) node[anchor=south]{$\mu$};

  \draw[->,thick,purple!70] (5,0) -- (5,1.9);
  \node[anchor=west,purple!70] at (6,1.9) {$A_D(\kappa)$};

  \draw[blue!60,thick,dashed] (5,0) -- ++(70:2) arc (70:110:2) -- cycle;
  \node[blue!60] at (4.3,2.1) {$\kappa$};

  \draw[->,thick,gray] (5,0) -- (7,0) node[anchor=west,gray]{$\mu_{\perp}$};

  \draw[->,thick] (5,0) -- ({5 + 1.6*cos(30)},{0 + 1.6*sin(30)}) node[anchor=west]{$x$};

  \draw[dashed] ({5 + 1.6*cos(30)},{0 + 1.6*sin(30)}) -- (5,{0 + 1.6*sin(30)});
  \draw[dashed] ({5 + 1.6*cos(30)},{0 + 1.6*sin(30)}) -- ({5 + 1.6*cos(30)},0);

  \draw[<->,purple!70,thick] (3.3,2.15) -- (3.3,2.45) node[anchor=south,purple!70] {$\sigma_{\parallel}$};
  \draw[<->,purple!70,thick] (2.9,2.3) -- (3.7,2.3) node[anchor=west,purple!70] {$\sigma_{\perp}$};

  \draw (10,0) circle (2);

  \draw[->,thick,blue!70] (10,0) -- (10,2) node[anchor=south]{$\mu$};

  \draw[-,thick,purple!70] (10,0) -- (10,{0 + 2*sin(15) + 0.12});
  \node[anchor=west,purple!70] at (12,{0 + 2*sin(15) + 0.12}) {$A_D(\kappa)$};

  \draw[blue!60,thick,dashed] (10,0) -- ++(15:2) arc (15:165:2) -- cycle;
  \node[blue!60,anchor=east] at (8.,1.1) {$\kappa$};

  \draw[->,thick,gray] (10,0) -- (12,0) node[anchor=west,gray]{$\mu_{\perp}$};

  \draw[<->,purple!70,thick]
    ({10 - 0.8*2*cos(15)},{0 + 2*sin(15) + 0.12}) -- ({10 + 0.8*2*cos(15)},{0 + 2*sin(15) + 0.12});
  \draw[<->,purple!70,thick]
    (10,{0 + 2*sin(15) - 0.2 + 0.12}) -- (10,{0 + 2*sin(15) + 0.2 + 0.12});
  \node[anchor=west,purple!70] at ({10 + 0.82*2*cos(15) - 0.4},{0 + 2*sin(15) + 0.22 + 0.12}) {$\sigma_{\perp}$};
  \node[anchor=west,purple!70] at (9.95,{0 + 2*sin(15) + 0.3 + 0.12}) {$\sigma_{\parallel}$};
\end{tikzpicture}
\caption{Geometric view of directional weight uncertainty on the sphere. Left: vMF posterior with a cap around $\mu$ and sample directions (gray). Middle: small cap with mean shrinkage $A_D(\kappa)\,\mu$ and variance decomposition $\{\sigma_{\parallel},\sigma_{\perp}\}$. Right: broad cap near the equator with dominant perpendicular variability.}
\label{fig:tikz_vmf_sphere}
\end{figure}

\section{From directional posteriors to activation variance}
\label{sec:variance}
We summarize exact identities and a compact approximation linking vMF concentration $\kappa$ to the variance of $u=w^\top x$. Establishing this link lets us replace the intractable concentration parameter with a practical, optimizable activation-noise parameter that integrates cleanly into training.

\subsection{Exact moments and spherical average}
For $w\sim\mathrm{vMF}(\mu,\kappa)$, $\mathbb E[w]=A_D(\kappa)\,\mu$ and $\mathrm{Cov}(w)=\sigma_{\parallel}^2\,\mu\mu^\top+\sigma_{\perp}^2(I-\mu\mu^\top)$ with coefficients given in terms of Bessel ratios \citep{MardiaJupp2000}. Averaging $\sigma_u^2(x;\kappa)=x^\top\mathrm{Cov}(w)x$ over $x$ uniformly on the sphere of radius $\|x\|$ yields
\begin{equation}
\overline{\sigma_u^2}(\kappa)=\frac{\|x\|^2}{D}\,\mathrm{Tr}\bigl(\mathrm{Cov}(w)\bigr)=\frac{\|x\|^2}{D}\bigl(1-A_D(\kappa)^2\bigr).\label{eq:spherical_average}
\end{equation}
With $\|x\|^2\approx D$, this becomes $\overline{\sigma_u^2}(\kappa)=1-A_D(\kappa)^2$.

\subsection{Compact interpolant}
For optimization we use a simple, dimension-aware interpolant that matches both limits:
\begin{equation}
\sigma_u^2(\kappa)\;\approx\;\frac{D}{\kappa+D},\label{eq:simple_interpolant}
\end{equation}
so $\sigma_u^2\to1$ for $\kappa\to0$ and $\sigma_u^2\sim D/\kappa$ for $\kappa\gg D$. This closely tracks the spherical-average identity in~\eqref{eq:spherical_average} while avoiding special functions.

\subsection{Effective noise and regime inversions}
Defining $\sigma_{\mathrm{eff}}=\sigma_u/A_D(\kappa)$ (Section~\ref{sec:geometry}) provides a direct post-normalization noise parameter. Two useful inversions are
\begin{equation}
\kappa\;\approx\;\frac{D}{\sigma_{\mathrm{eff}}^2}\quad(\sigma_{\mathrm{eff}}\ll1),\qquad\kappa\;\approx\;\frac{D}{\sigma_{\mathrm{eff}}}\quad(\sigma_{\mathrm{eff}}\gg1),\label{eq:regime_inversions}
\end{equation}
capturing the tight- and broad-cap regimes, respectively. We use $\sigma_{\mathrm{eff}}$ as the optimized variable and evaluate the KL directly in $(\sigma_{\mathrm{eff}},D)$ in the next section.

\subsection{Quick-reference tables}
\begin{table}[t]
\centering
\caption{Regimes and leading relations.}
\label{tab:regimes}
\begingroup
\renewcommand{\arraystretch}{1.25}
\begin{tabular}{lcccc}
\hline
Regime & Relation $\kappa(\sigma_{\mathrm{eff}})$ & $\sigma_u^2$ & $\sigma_{\mathrm{eff}}$ & KL (leading) \\
\hline
Tight $\kappa\gg D$ & $\displaystyle \kappa\approx\frac{D}{\sigma_{\mathrm{eff}}^2}$ & $\displaystyle \frac{D}{\kappa}$ & $\displaystyle \sqrt{\frac{D}{\kappa}}$ & $\displaystyle \frac{D-1}{2}\,\log\frac{D}{\sigma_{\mathrm{eff}}^2}$ \\
Broad $\kappa\ll D$ & $\displaystyle \kappa\approx\frac{D}{\sigma_{\mathrm{eff}}}$ & $\displaystyle 1$ & $\displaystyle \frac{D}{\kappa}$ & $\displaystyle \frac{D}{2}\,\sigma_{\mathrm{eff}}^{-2}$ \\
Unified & $-$ & $\displaystyle \frac{D}{\kappa+D}$ & $\displaystyle \frac{\sqrt{D/(\kappa+D)}}{A_D(\kappa)}$ & $\displaystyle \frac{D-1}{2}\,\log\Bigl(1+\frac{D}{D-1}\,\sigma_{\mathrm{eff}}^{-2}\Bigr)$ \\
\hline
\end{tabular}
\endgroup
\end{table}

\begin{table}[t]
\centering
\caption{Key formulas used in training.}
\label{tab:key_formulas}
\begingroup
\renewcommand{\arraystretch}{1.25}
\begin{tabular}{ll}
\hline
Quantity & Expression \\
\hline
Activation variance & $\displaystyle \sigma_u^2(\kappa)\approx\frac{D}{\kappa+D}$ \\
Mean resultant length & $\displaystyle A_D(\kappa)=\frac{I_{D/2}(\kappa)}{I_{D/2-1}(\kappa)}$ \\
Effective noise & $\displaystyle \sigma_{\mathrm{eff}}=\sigma_u/A_D(\kappa)$ \\
KL (training) & $\displaystyle \mathrm{KL}_{\mathrm{approx}}(\sigma_{\mathrm{eff}},D)=\frac{D-1}{2}\,\log\Bigl(1+\frac{D}{D-1}\,\sigma_{\mathrm{eff}}^{-2}\Bigr)$ \\
Objective & $\displaystyle \mathcal L=\mathrm{NLL}+\beta\sum_{\ell} M_\ell\,\mathrm{KL}_{\mathrm{approx}}\bigl(\sigma_{\mathrm{eff},\ell},D_\ell\bigr)$ \\
\hline
\end{tabular}
\endgroup
\end{table}

\section{KL in terms of $\sigma_{\mathrm{eff}}$ and the variational objective}
\label{sec:kl_objective}
The exact KL between a vMF posterior and the uniform prior on the sphere is
\begin{equation}
\mathrm{KL}(\kappa)=\kappa\,A_D(\kappa)+\log C_D(\kappa)-\log \mathrm{Area}(\mathbb S^{D-1}),
\end{equation}
which is straightforward to evaluate numerically but is computationally inconvenient for gradient-based optimization due to special functions, and it obscures the direct relationship between the optimization variable and the KL penalty. We therefore use a closed-form approximation written directly in terms of the optimized variable $\sigma_{\mathrm{eff}}$ and the input dimension $D$:
\begin{equation}
\mathrm{KL}_{\mathrm{approx}}(\sigma_{\mathrm{eff}},D)\;=\;\frac{D-1}{2}\,\log\Bigl(1+\frac{D}{D-1}\,\sigma_{\mathrm{eff}}^{-2}\Bigr).\label{eq:kl_sigma}
\end{equation}
This form matches the vMF-vs-uniform KL asymptotics (logarithmic for tight posteriors, quadratic in $\sigma_{\mathrm{eff}}^{-2}$ for broad posteriors) \citep{Sra2016}; it is numerically stable and \emph{dimension-aware} through the $(D-1)/2$ factor. The leading factor properly scales the regularization to the layer's capacity, preventing the KL from vanishing or exploding in wide or narrow layers.

\paragraph{Training objective.} For a model with a set of normalized layers $\mathcal L$, each having input dimension $D_\ell$, multiplicity $M_\ell$ (e.g., number of output channels), and a single learned scalar $\sigma_{\mathrm{eff},\ell}>0$, we minimize
\begin{equation}
\mathcal L\;=\;\mathrm{NLL}(\text{data})\;+\;\beta\,\sum_{\ell\in\mathcal L} M_\ell\,\mathrm{KL}_{\mathrm{approx}}\bigl(\sigma_{\mathrm{eff},\ell},\,D_\ell\bigr),\label{eq:total_objective}
\end{equation}
with optional KL warm-up via $\beta\in[0,1]$. The KL scales with the layer's input dimensionality $D_\ell$.

\paragraph{Forward pass (training).} For a normalized linear/convolutional block with unit-norm weights $\tilde w$, compute $u=\tilde w^\top x$, apply BN, and add Gaussian noise: $y=\mathrm{BN}(u)+\sigma_{\mathrm{eff}}\,\varepsilon$, $\varepsilon\sim\mathcal N(0,1)$. At evaluation, set $y=\mathrm{BN}(u)$ (or average a few stochastic passes for predictive uncertainty). Gradients in $\sigma_{\mathrm{eff}}$ use the reparameterization trick \citep{Kingma2014autoencoding} and the analytic KL in~\eqref{eq:kl_sigma}.

\section{Experiments}
\label{sec:experiments}
We validate the framework's principles and core mechanics rather than pursue exhaustive benchmarks: (i) Do the variance and KL approximations match ground truth across regimes? (ii) Does $\sigma_{\mathrm{eff}}$ recover task difficulty in controlled settings? (iii) Does the method improve calibration on CIFAR-10 while maintaining accuracy?

\subsection{Experimental setup}
\textbf{Synthetic (student--teacher).} Linear teacher with normalized weights; the student minimizes NLL+$\beta$KL using $\mathrm{KL}_{\mathrm{approx}}(\sigma_{\mathrm{eff}},D)$. We sweep observation noise, input dimension $D$, and sample size $N$.

\textbf{CIFAR-10.} VGG16 with BatchNorm (affine=False). We insert one post-normalization additive-noise parameter $\sigma_{\mathrm{eff}}$ per normalized layer and apply a KL warm-up. Metrics: accuracy, negative log-likelihood (NLL), and Expected Calibration Error (ECE) \citep{Guo2017calibration}. Training details (optimizer, schedules, and architecture) follow standard practice and are given in the appendix.

\subsection{Validation: synthetic theory checks}
\begin{figure}[t]
\centering
\includegraphics[width=0.9\textwidth]{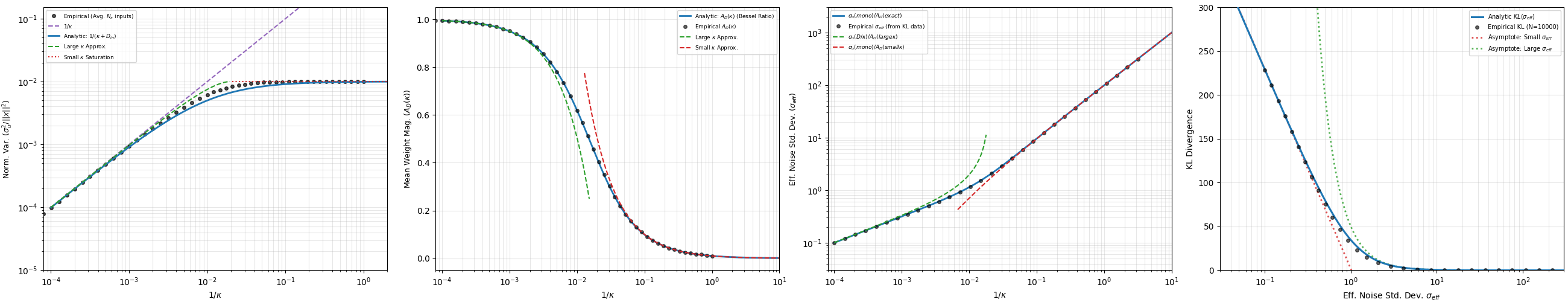}
\caption{Validation of theoretical components at $D=100$. (A) Activation variance: Monte Carlo vs. interpolant $D/(\kappa+D)$. (B) Mean resultant length $A_D(\kappa)$ (Bessel ratio) vs. MC. (C) Effective post-normalization noise $\sigma_{\mathrm{eff}}=\sigma_u/A_D(\kappa)$ alignment. (D) KL: numerical vMF-vs-uniform vs. $\mathrm{KL}_{\mathrm{approx}}(\sigma_{\mathrm{eff}},D)$ with asymptotes.}
\label{fig:theory_summary}
\end{figure}
Across a wide $\kappa$ range, Monte Carlo measurements of $\sigma_u^2$ match $D/(\kappa+D)$; the closed-form KL in~\eqref{eq:kl_sigma} tracks the numerically evaluated vMF KL, including both asymptotic regimes.

\subsection{Student-teacher parameter recovery}
\begin{figure}[t]
\centering
\includegraphics[width=\textwidth]{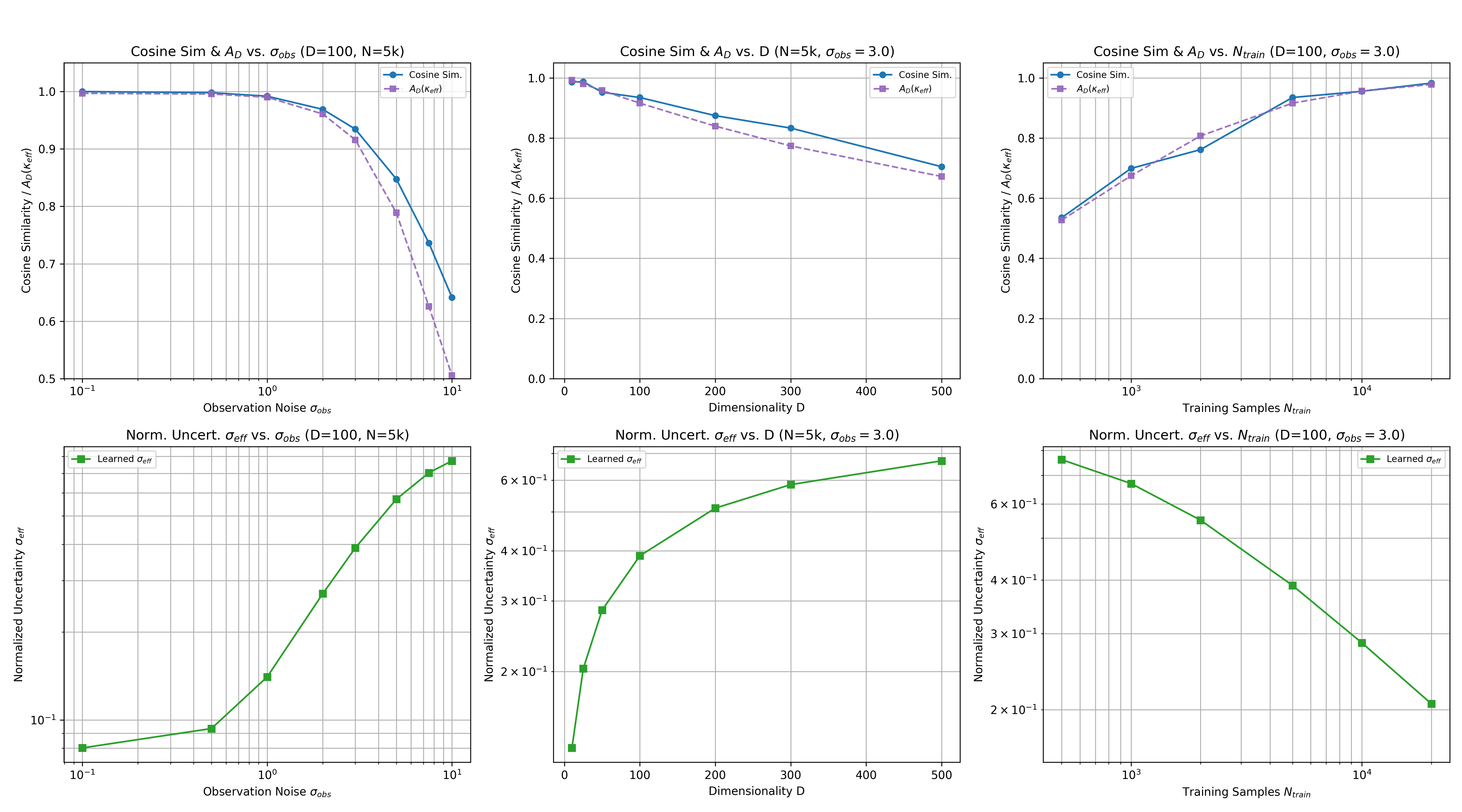}
\caption{Student-teacher recovery. Top: cosine similarity and inferred $A_D(\kappa_{\mathrm{eff}})$ track each other across sweeps. Bottom: learned $\sigma_{\mathrm{eff}}$ adjusts coherently with observation noise, dimension $D$, and sample size $N$.}
\label{fig:student_teacher}
\end{figure}
The learned $\sigma_{\mathrm{eff}}$ increases with observation noise, grows with $D$ at fixed $N$, and decreases with more data, while direction recovery aligns with $A_D(\kappa_{\mathrm{eff}})$ inferred from $\sigma_{\mathrm{eff}}$.

\subsection{CIFAR-10 calibration and accuracy}
\begin{figure}[t]
\centering
\includegraphics[width=\textwidth,clip,trim=0 0 0 24]{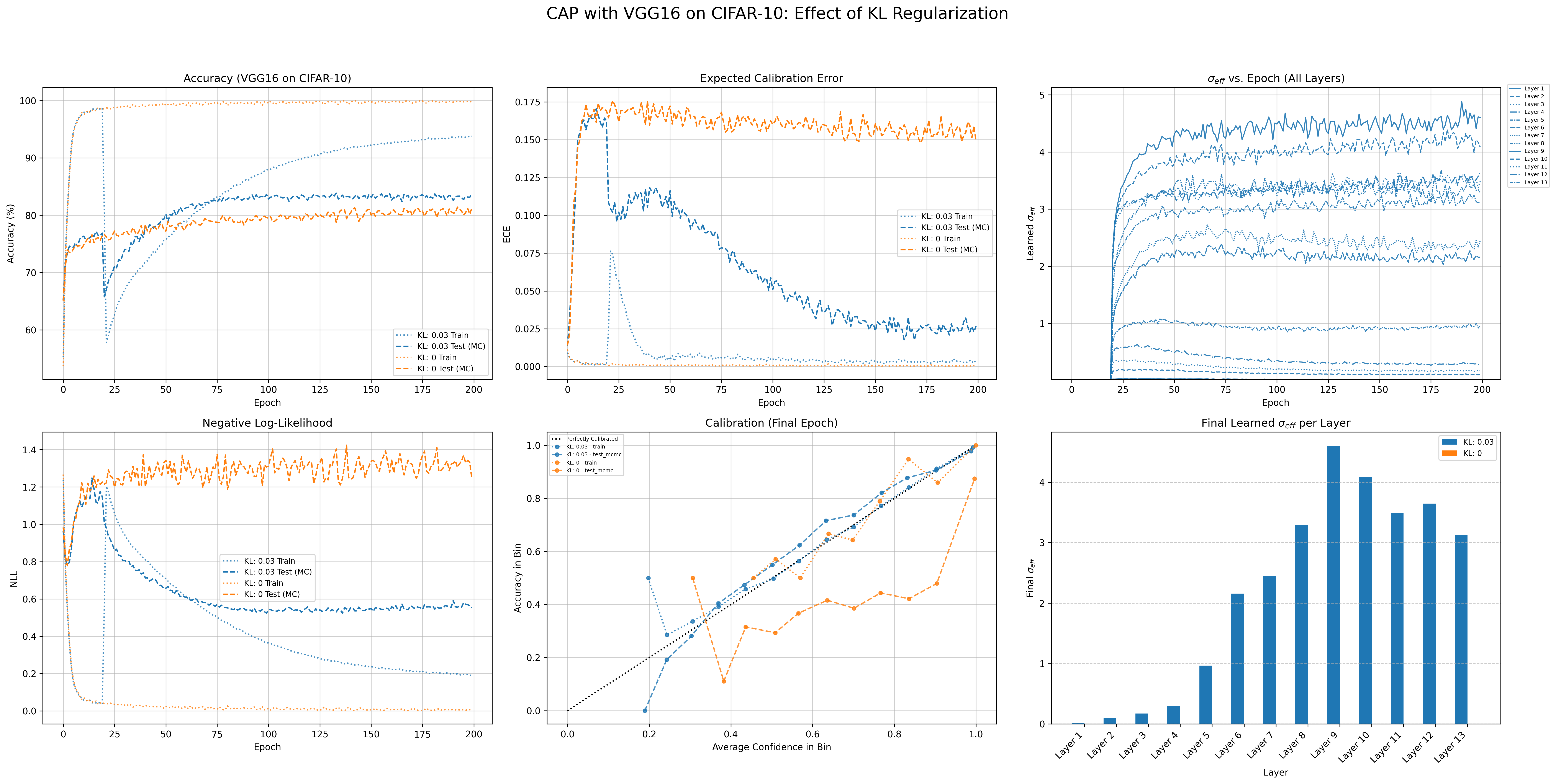}
\caption{CIFAR-10 (VGG16+BN). Left-to-right/top-to-bottom: accuracy, NLL, calibration curves, ECE, training dynamics of $\sigma_{\mathrm{eff}}$, and per-layer $\sigma_{\mathrm{eff}}$ profile. The proposed geometry-aligned method reduces ECE by $\approx$5.6$\times$ while maintaining competitive accuracy and lower NLL.}
\label{fig:cifar}
\end{figure}
Compared to a baseline without KL, our method substantially improves calibration (ECE $\downarrow$) with similar or better accuracy and lower NLL. The learned per-layer $\sigma_{\mathrm{eff}}$ profile is structured (larger mid-depth), providing interpretable uncertainty with minimal overhead.

\begin{figure}[t]
\centering
\begin{minipage}[t]{0.49\textwidth}
  \centering
  \includegraphics[width=\linewidth]{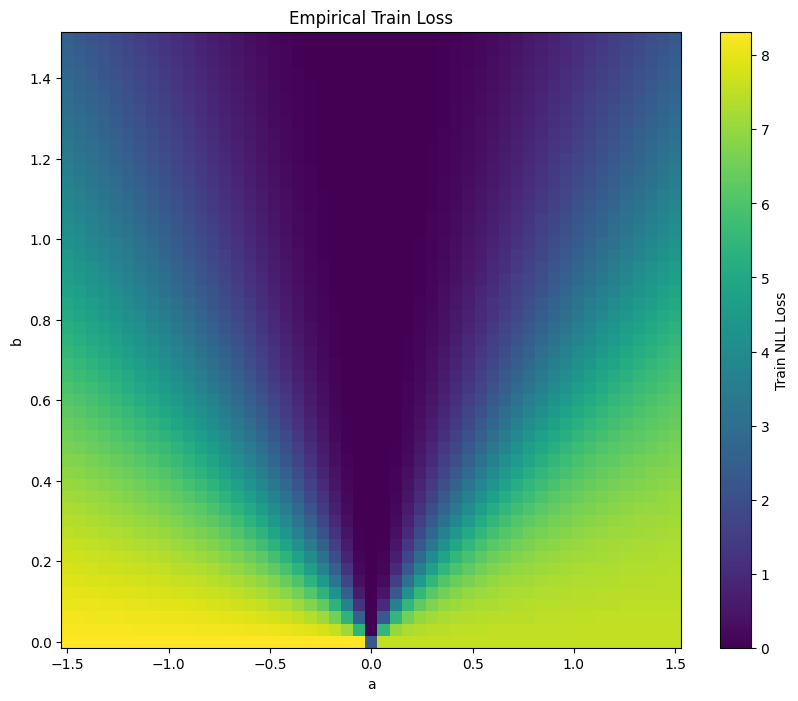}
  \vspace{0.2em}
  \small Empirical 1D (train): $w' = y\,\hat w + x\,\hat r$.
\end{minipage}
\hfill
\begin{minipage}[t]{0.49\textwidth}
  \centering
  \includegraphics[width=\linewidth]{figures/loss_landscape_cap_e200_2D_51x51_20250824-150744/landscape_features_0_vgg16_cap_effkl_burnin20_e200_twoD_Train_Loss.png}
  \vspace{0.2em}
  \small Empirical 2D (train): $w' = \hat w + a\,\hat r_1 + b\,\hat r_2$.
\end{minipage}

\vspace{0.6em}

\begin{minipage}[t]{0.49\textwidth}
  \centering
  \includegraphics[width=\linewidth]{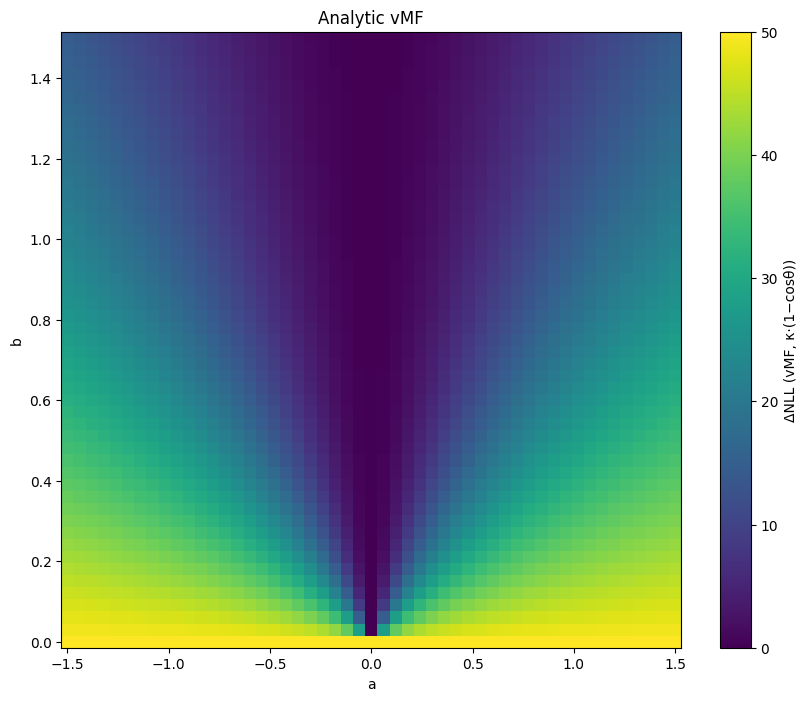}
  \vspace{0.2em}
  \small Analytic vMF.
\end{minipage}
\hfill
\begin{minipage}[t]{0.49\textwidth}
  \centering
  \includegraphics[width=\linewidth]{figures/paper_figures/vmf_2d_layer_features0_k50.png}
  \vspace{0.2em}
  \small Analytic vMF.
\end{minipage}

\caption{Empirical train-loss landscapes (top) and analytic vMF predictions (bottom) around the first convolutional layer. The analytic forms use $\Delta\mathcal L = \kappa(1-\cos\theta)$ with tangent directions orthogonal to the trained direction. The shapes match: bowl-like 1D cross-section and approximately radial 2D level sets, supporting a directional uncertainty model.}
\label{fig:empirical_vs_vmf_landscapes}
\end{figure}

\paragraph{Loss landscapes around a trained model.} To probe local invariances and sensitivity, we visualize the training loss landscape around the first convolutional layer by overwriting its weights with normalized perturbations while keeping the network in train mode (BN uses batch statistics; noise modules set to eval so no noise is injected). We use two parameterizations with coordinates $(a,b)$: (1) 1D: $w' = y\,\hat w + x\,\hat r$ (plotted with axes renamed to $a,b$), and (2) 2D: $w' = \hat w + a\,\hat r_1 + b\,\hat r_2$ with $\hat r_1\perp \hat r_2$. The center corresponds to the unmodified model. Figure~\ref{fig:empirical_vs_vmf_landscapes} (top) shows the empirical train-loss maps; the bottom row plots the analytic vMF prediction $\Delta\mathcal L=\kappa(1-\cos\theta)$ on the same coordinate systems, highlighting the same bowl-like 1D profile and near-radial 2D level sets.

\subsection{Implementation details}
We use per-layer $\log\sigma_{\mathrm{eff}}$ parameters with softplus/exp for positivity, unit-norm weights (or weight normalization with fixed gain), and KL warm-up. Inference is deterministic (noise off) or Monte Carlo with few samples; the KL adds negligible cost.

\section{Discussion and Conclusion}
We presented a variational framework for BNNs built from a simple but fundamental insight: the variational posterior should respect the geometry induced by the network's architecture. By modeling uncertainty in weight directions—the primary degree of freedom in normalized networks—we arrived at a practical and precise method. High-dimensional spherical geometry yields a compact activation-variance interpolant and a closed-form, dimension-aware KL in the natural optimization variable $\sigma_{\mathrm{eff}}$. The resulting “one scalar per layer” design integrates with standard architectures, improves calibration, and remains efficient.

\textbf{Limitations.} (i) Small input dimensions (e.g., early conv layers) weaken high-dimensional asymptotics; while the KL remains stable, approximations may be looser. (ii) Early in training, a stationary posterior may be a poor description of rapidly changing weights; warm-up mitigates but does not eliminate this. (iii) We focus on classification with post-normalization additive noise; regression or non-BN pipelines may require minor adaptations.

\textbf{Future work.} (i) Adaptive KL schedules tied to training dynamics; (ii) extensions to other normalizations and group-wise/shared $\sigma_{\mathrm{eff}}$; (iii) layer-wise or data-dependent priors on the sphere; (iv) combining with ensembling or SWA for stronger uncertainty under distribution shift. Ultimately, robust and practical Bayesian deep learning may come not from more complex posteriors, but from posteriors more thoughtfully aligned with modern network design.

\bibliography{references}

\begin{thebibliography}{22}
\providecommand{\natexlab}[1]{#1}
\providecommand{\url}[1]{\texttt{#1}}
\expandafter\ifx\csname urlstyle\endcsname\relax
  \providecommand{\doi}[1]{doi: #1}\else
  \providecommand{\doi}{doi: \begingroup \urlstyle{rm}\Url}\fi

\bibitem[Banerjee et~al.(2005)Banerjee, Dhillon, Ghosh, and
  Sra]{Banerjee2005clustering}
Anindya Banerjee, Inderjit~S. Dhillon, Joydeep Ghosh, and Suvrit Sra.
\newblock Clustering on the unit hypersphere using von mises-fisher
  distributions.
\newblock \emph{Journal of Machine Learning Research}, 6:\penalty0 1345--1382,
  2005.

\bibitem[Bishop(2006)]{Bishop2006}
Christopher~M. Bishop.
\newblock \emph{Pattern Recognition and Machine Learning}.
\newblock Springer, 2006.

\bibitem[Davidson et~al.(2018)Davidson, Falorsi, De~Cao, Kipf, and
  Tomczak]{davidson2018hyperspherical}
Tim~R. Davidson, Luca Falorsi, Nicola De~Cao, Thomas Kipf, and Jakub~M.
  Tomczak.
\newblock Hyperspherical variational auto-encoders.
\newblock In \emph{Advances in Neural Information Processing Systems 31}, pp.\
  330--340. Curran Associates, Inc., 2018.

\bibitem[Farquhar et~al.(2020)Farquhar, Osborne, and Gal]{farquhar2020radial}
Sebastian Farquhar, Michael~A Osborne, and Yarin Gal.
\newblock Radial bayesian neural networks: Beyond discrete support in
  large-scale bayesian deep learning.
\newblock In \emph{International Conference on Artificial Intelligence and
  Statistics}, pp.\  1352--1362. PMLR, 2020.

\bibitem[Gal \& Ghahramani(2016)Gal and Ghahramani]{Gal2016dropout}
Yarin Gal and Zoubin Ghahramani.
\newblock Dropout as a bayesian approximation: Representing model uncertainty
  in deep learning.
\newblock In \emph{Proceedings of The 33rd International Conference on Machine
  Learning}, volume~48 of \emph{PMLR}, pp.\  1050--1059. PMLR, 2016.

\bibitem[Ghoshal \& Tucker(2021)Ghoshal and Tucker]{ghoshal2021hyperspherical}
Biraja Ghoshal and Allan Tucker.
\newblock Hyperspherical weight uncertainty in neural networks.
\newblock In \emph{Advances in Intelligent Data Analysis XIX}, pp.\  3--14.
  Springer, 2021.

\bibitem[Graves(2011)]{Graves2011practicalvi}
Alex Graves.
\newblock Practical variational inference for neural networks.
\newblock In \emph{Advances in Neural Information Processing Systems 24}, pp.\
  2348--2356. Curran Associates, Inc., 2011.

\bibitem[Guo et~al.(2017)Guo, Pleiss, Sun, and Weinberger]{Guo2017calibration}
Chuan Guo, Geoff Pleiss, Yu~Sun, and Kilian~Q Weinberger.
\newblock On calibration of modern neural networks.
\newblock \emph{International Conference on Machine Learning}, pp.\
  1321--1330, 2017.

\bibitem[Ioffe \& Szegedy(2015)Ioffe and Szegedy]{Ioffe2015batchnorm}
S.~Ioffe and C.~Szegedy.
\newblock Batch normalization: Accelerating deep network training by reducing
  internal covariate shift.
\newblock In \emph{Proceedings of the International Conference on Machine
  Learning}, 2015.

\bibitem[Izmailov et~al.(2021)Izmailov, Vikram, Hoffman, and
  Wilson]{Izmailov2021bayesian}
Pavel Izmailov, Sharad Vikram, Matthew~D Hoffman, and Andrew~Gordon Wilson.
\newblock What are bayesian neural network posteriors really like?
\newblock \emph{International Conference on Machine Learning}, pp.\
  4629--4640, 2021.

\bibitem[Jordan et~al.(1999)Jordan, Ghahramani, Jaakkola, and Saul]{Jordan1999}
Michael~I. Jordan, Zoubin Ghahramani, Tommi~S. Jaakkola, and Lawrence~K. Saul.
\newblock An introduction to variational methods for graphical models.
\newblock \emph{Machine Learning}, 37\penalty0 (2):\penalty0 183--233, 1999.

\bibitem[Kingma \& Welling(2014)Kingma and Welling]{Kingma2014autoencoding}
Diederik~P. Kingma and Max Welling.
\newblock Auto-encoding variational bayes.
\newblock In \emph{Proceedings of the 2nd International Conference on Learning
  Representations, {ICLR} 2014, Banff, AB, Canada, April 14-16, 2014,
  Conference Track Proceedings}, 2014.

\bibitem[Kingma et~al.(2015)Kingma, Salimans, and
  Welling]{Kingma2015variational}
Diederik~P. Kingma, Tim Salimans, and Max Welling.
\newblock Variational dropout and the local reparameterization trick.
\newblock In \emph{Advances in Neural Information Processing Systems 28}, pp.\
  2575--2583. Curran Associates, Inc., 2015.

\bibitem[Mardia \& Jupp(2000)Mardia and Jupp]{MardiaJupp2000}
Kanti~V. Mardia and Peter~E. Jupp.
\newblock \emph{Directional Statistics}.
\newblock John Wiley \string& Sons, 2000.

\bibitem[Meng et~al.(2019)Meng, Zheng, Zhang, Chen, Ma, and Liu]{Meng2019gsgd}
Qi~Meng, Shuxin Zheng, Huishuai Zhang, Wei Chen, Zhi-Ming Ma, and Tie-Yan Liu.
\newblock {G-SGD}: Optimizing {ReLU} neural networks in its positively
  scale-invariant space.
\newblock In \emph{International Conference on Learning Representations
  (ICLR)}, 2019.

\bibitem[Molchanov et~al.(2017)Molchanov, Ashukha, and
  Vetrov]{Molchanov2017variational}
Dmitry Molchanov, Arsenii Ashukha, and Dmitry Vetrov.
\newblock Variational dropout sparsifies deep neural networks.
\newblock In \emph{Proceedings of the 34th International Conference on Machine
  Learning, {ICML} 2017, Sydney, NSW, Australia, 6-11 August 2017}, volume~70
  of \emph{Proceedings of Machine Learning Research}, pp.\  2498--2507. PMLR,
  2017.

\bibitem[Mukhoti et~al.(2020)Mukhoti, Kulharia, Sanyal, Golodetz, Torr, and
  Dokania]{Mukhoti2020calibrating}
Jishnu Mukhoti, Viveka Kulharia, Amartya Sanyal, Stuart Golodetz, Philip Torr,
  and Puneet Dokania.
\newblock Calibrating deep neural networks using focal loss.
\newblock \emph{Advances in Neural Information Processing Systems},
  33:\penalty0 15744--15755, 2020.

\bibitem[Neyshabur et~al.(2015)Neyshabur, Tomioka, and
  Srebro]{Neyshabur2015normbased}
Behnam Neyshabur, Ryota Tomioka, and Nathan Srebro.
\newblock Norm-based capacity control in neural networks.
\newblock In \emph{Proceedings of The 28th Conference on Learning Theory},
  volume~40 of \emph{Proceedings of Machine Learning Research}, pp.\
  1376--1401. PMLR, 2015.

\bibitem[Oh et~al.(2020)Oh, Adamczewski, and Park]{oh2020radial}
Changyong Oh, Kamil Adamczewski, and Mijung Park.
\newblock Radial and directional posteriors for bayesian deep learning.
\newblock \emph{Proceedings of the AAAI Conference on Artificial Intelligence},
  34\penalty0 (04):\penalty0 5298--5305, 2020.

\bibitem[Salimans \& Kingma(2016)Salimans and Kingma]{Salimans2016weightnorm}
T.~Salimans and D.~P. Kingma.
\newblock Weight normalization: A simple reparameterization to accelerate
  training of deep neural networks.
\newblock In \emph{Advances in Neural Information Processing Systems}, 2016.

\bibitem[Sra(2016)]{Sra2016}
Sanjoy Sra.
\newblock Directional statistics in machine learning: A brief review.
\newblock \emph{arXiv preprint arXiv:1605.00316}, 2016.

\bibitem[Yang(2022)]{Yang2022tensorprogramsV}
Greg Yang.
\newblock Tensor programs v: Tuning large neural networks via
  $\mu$-parametrization.
\newblock \emph{arXiv preprint arXiv:2203.03466}, 2022.

\end{thebibliography}
\bibliographystyle{iclr2024/iclr2024_conference}

\appendix

\section{Implementation details}
\label{app:implementation}
\paragraph{Parameterization and forward.} Each layer maintains a single positive scalar $\sigma_{\mathrm{eff}}$ (we optimize $\rho$ with $\sigma_{\mathrm{eff}}=\mathrm{softplus}(\rho)$ or $\exp(\rho)$). Given unit-norm weights $\tilde w$ and input $x$, compute $u=\tilde w^\top x$, apply BN (affine=False), and output $y=\mathrm{BN}(u)+\sigma_{\mathrm{eff}}\,\varepsilon$, $\varepsilon\sim\mathcal N(0,1)$. At evaluation, use $y=\mathrm{BN}(u)$ (or MC sampling for uncertainty).

\paragraph{Objective and scaling.} The training loss is $\mathrm{NLL}+\beta\sum_\ell M_\ell\,\mathrm{KL}_{\mathrm{approx}}(\sigma_{\mathrm{eff},\ell},D_\ell)$ with multiplicity $M_\ell$ (e.g., output channels) and input dimension $D_\ell$. We use a short KL warm-up from 0 to 1 to stabilize early training.

\paragraph{Synthetic student-teacher.} Teacher weights are unit-normalized; data are generated with Gaussian observation noise. The student minimizes NLL+$\beta$KL with Adam (lr $\approx 1\mathrm{e}{-3}$), sweeping observation noise, input dimension $D$, and sample size $N$.

\paragraph{CIFAR-10.} VGG16 with BN (affine=False). We insert a post-normalization additive-noise parameter after each BN, optimize base parameters with lr $1\mathrm{e}{-4}$ and the noise parameters with a higher lr (e.g., $2\mathrm{e}{-2}$), batch size 256, 200 epochs, and a 20-epoch KL warm-up. Metrics: accuracy, NLL, and ECE (15 bins). We normalize inputs with dataset mean/std.

\paragraph{Loss landscapes.} To probe local geometry, we overwrite a chosen layer's weights with normalized perturbations: 1D $w'= y\,\hat w + x\,\hat r$, $x\in[-1.5,1.5],\ y\in[0,1.5]$; 2D $w'=\hat w + a\,\hat r_1 + b\,\hat r_2$, $(a,b)\in[-1.5,1.5]^2$. The network remains in train mode (BN uses batch stats), noise modules are set to eval.

\section{vMF moments and covariance}
\label{app:vmf_moments}
The vMF density on $\mathbb S^{D-1}$ is $q(w\mid\mu,\kappa)=C_D(\kappa)\exp(\kappa\mu^\top w)$ with
\begin{equation}
C_D(\kappa)=\frac{\kappa^{D/2-1}}{(2\pi)^{D/2} I_{D/2-1}(\kappa)},\qquad A_D(\kappa):=\frac{I_{D/2}(\kappa)}{I_{D/2-1}(\kappa)}.
\end{equation}
Moments: $\mathbb E[w]=A_D(\kappa)\,\mu$ and $\mathrm{Cov}(w)=\sigma_{\parallel}^2\,\mu\mu^\top+\sigma_{\perp}^2(I-\mu\mu^\top)$, where closed forms for $\sigma_{\parallel}^2,\sigma_{\perp}^2$ follow from Bessel identities \citep{MardiaJupp2000}.

\section{Spherical-average identity}
\label{app:spherical_avg}
For any positive semidefinite $S\in\mathbb R^{D\times D}$ and $x$ uniformly distributed on the sphere of radius $r$, $\mathbb E[ x^\top S x ] = r^2\,\mathrm{Tr}(S)/D$. With $S=\mathrm{Cov}(w)$ and $\|x\|^2\approx D$,
\begin{equation}
\overline{\sigma_u^2}(\kappa)=\frac{\|x\|^2}{D}\,\mathrm{Tr}\bigl(\mathrm{Cov}(w)\bigr)=\frac{\|x\|^2}{D}\bigl(1-A_D(\kappa)^2\bigr)\approx 1-A_D(\kappa)^2.
\end{equation}

\section{Compact interpolant justification}
\label{app:interpolant}
We use $f(\kappa):=D/(\kappa+D)$ to approximate $\sigma_u^2$. It satisfies $f(0)=1$ (broad cap) and $f(\kappa)\sim D/\kappa$ (tight cap), matching the leading terms of $1-A_D(\kappa)^2\sim (D-1)/\kappa$ up to a coefficient difference of order $1/D$. Empirically, the relative error is small over practical $(D,\kappa)$ ranges.

\section{Asymptotics of $A_D(\kappa)$}
\label{app:ad_asymptotics}
Standard expansions (via Bessel asymptotics):
\begin{align}
\text{Small }\kappa:&\quad A_D(\kappa)=\frac{\kappa}{D}-\frac{\kappa^3}{D(D+2)}+O(\kappa^5),\\
\text{Large }\kappa:&\quad A_D(\kappa)=1-\frac{D-1}{2\kappa}+\frac{(D-1)(D-3)}{8\kappa^2}+O(\kappa^{-3}).
\end{align}
Substituting into $1-A_D(\kappa)^2$ recovers the limiting behaviors in the main text.

\section{KL asymptotics and $\mathrm{KL}_{\mathrm{approx}}$}
\label{app:kl}
Let $Z(\kappa):=C_D(\kappa)^{-1}=\int_{\mathbb S^{D-1}}\exp(\kappa\mu^\top w)\,dw$. The vMF-vs-uniform KL is $\kappa A_D(\kappa)-\log Z(\kappa)-\log \mathrm{Area}(\mathbb S^{D-1})$. For large $\kappa$, Laplace's method yields $\log Z(\kappa)=\kappa+\tfrac{D-1}{2}\log(2\pi/\kappa)+O(1)$ and thus $\mathrm{KL}(\kappa)\sim \tfrac{D-1}{2}\log\kappa+O(1)$ \citep{Sra2016}. For small $\kappa$, a Taylor expansion combined with the spherical-average identity implies a quadratic leading term when expressed in $\sigma_{\mathrm{eff}}^{-2}$. The closed form
\begin{equation}
\mathrm{KL}_{\mathrm{approx}}(\sigma_{\mathrm{eff}},D)=\frac{D-1}{2}\,\log\Bigl(1+\frac{D}{D-1}\,\sigma_{\mathrm{eff}}^{-2}\Bigr)
\end{equation}
matches both limits with correct coefficients and avoids Bessel functions in the optimization path.

\section{Regime inversions}
\label{app:inversions}
Using $\sigma_u^2\approx D/(\kappa+D)$ and the asymptotics of $A_D(\kappa)$, we obtain
\begin{equation}
\kappa\approx\frac{D}{\sigma_{\mathrm{eff}}^2}\quad(\sigma_{\mathrm{eff}}\ll1),\qquad \kappa\approx\frac{D}{\sigma_{\mathrm{eff}}}\quad(\sigma_{\mathrm{eff}}\gg1),
\end{equation}
useful for diagnostics and interpretation; training optimizes $\sigma_{\mathrm{eff}}$ directly.

\end{document}